\documentclass[runningheads]{llncs}
\usepackage{graphicx}

\usepackage{xcolor}%定义了一些颜色  
\usepackage{colortbl,booktabs}%第二个包定义了几个*rule 
\usepackage{graphicx}
\usepackage{epstopdf}
\usepackage{blindtext, graphicx, mathrsfs, amsmath, amsfonts}
\usepackage{booktabs}
\usepackage{multirow, minipage-marginpar}
\usepackage{floatrow}
\usepackage{caption}
\usepackage{subfigure}
\usepackage[marginal]{footmisc}
\begin{document}
\title{Double-uncertainty Weighted Method for Semi-supervised Learning}%\thanks{Supported by organization x.}}
%\title{Contribution Title\thanks{Supported by organization x.}}
%
\titlerunning{Double-uncertainty Weighted Method for Semi-supervised Learning}
\author{Yixin Wang\inst{1,2,3} \and
Yao Zhang\inst{1,2,3} \and
Jiang Tian\inst{3}\and
Cheng Zhong\inst{3}\and
Zhongchao Shi\inst{3}\and
Yang Zhang\inst{4}\and
Zhiqiang He\inst{1,2,4}}%1{Yixin Wang}%2{Yao Zhang}%3{Jiang Tian}%4{Cheng Zhong}%5{Zhongchao Shi}%6{Yang Zhang}%7{Zhiqiang He}
\authorrunning{Y. Wang et al.}
\institute{Institute of Computing Technology, Chinese Academy of Sciences, Beijing, China  \and
University of Chinese Academy of Sciences, Beijing, China
\email{\{wangyixin19,zhangyao215\}@mails.ucas.ac.cn}\\ \and
AI Lab, Lenovo Research, Beijing, China\and
Lenovo Corporate Research \& Development, Lenovo Ltd., Beijing, China 
%\email{\{zhangyang20,hezq\}@lenovo.com}
}
\renewcommand{\thefootnote}{}
\maketitle              % typeset the header of the contribution
\footnote{Yang Zhang and Zhiqiang He---Equal contribution. This work was done when Yixin Wang and Yao Zhang were interns at AI Lab, Lenovo Research.}
\begin{abstract}
Though deep learning has achieved advanced performance recently, it remains a challenging task in the field of medical imaging, as obtaining reliable labeled training data is time-consuming and expensive.
In this paper, we propose a double-uncertainty weighted method for semi-supervised segmentation based on the teacher-student model. The teacher model provides guidance for the student model by penalizing their inconsistent prediction on both labeled and unlabeled data. We train the teacher model using Bayesian deep learning to obtain double-uncertainty, i.e. segmentation uncertainty and feature uncertainty. It is the first to extend segmentation uncertainty estimation to feature uncertainty, which reveals the capability to capture information among channels. A learnable uncertainty consistency loss is designed for the unsupervised learning process in an interactive manner between prediction and uncertainty. With no ground-truth for supervision, it can still incentivize more accurate teacher's predictions and facilitate the model to reduce uncertain estimations. Furthermore, our proposed double-uncertainty serves as a weight on each inconsistency penalty to balance and harmonize supervised and unsupervised training processes. We validate the proposed feature uncertainty and loss function through qualitative and quantitative analyses. Experimental results show that our method outperforms the state-of-the-art uncertainty-based semi-supervised methods on two public medical datasets.
%We present the first exploration of channel uncertainty as feature uncertainty based on MC dropout
\end{abstract}
\section{Introduction}
There are great progresses in medical image segmentation using deep learning, such as U-Net~\cite{unet} and V-Net~\cite{vnet}, in the past few years. The accuracy and robustness of these models heavily depend on the quantity and quality of the training data. However, it is well known that high quality annotated medical data is quite expensive due to
the domain knowledge prerequisite. As a result, semi-supervised learning has been recently explored to leverage unlabeled data for medical image segmentation~\cite{Bai2017,trainsformation,Few-shot,DAN}.
Teacher-student framework, a popular semi-supervised method, has been successfully applied to medical image segmentation tasks on some organs~\cite{Cui,Sedai,UAMT}. Concretely, the teacher model is a Temporal Ensemble~\cite{Temporal} of the current model with perturbations, which yields more accurate targets~\cite{MT}. The student model learns from the teacher model through penalizing inconsistent prediction between them, defined as consistency loss. However, with no ground-truth given for unlabeled training data, it is hard to judge whether the teacher model provides accurate prediction. 

To alleviate this problem, uncertainty measures are considered to be the optimal strategy, due to their capability to detect when and where the model is likely to make false predictions or the input is out-of-distribution. Recent explorations on uncertainty estimations include Bayesian uncertainty estimation via test-time dropout~\cite{MC} and network ensembling~\cite{Ensemble}. In the medical domain, Yu et al.~\cite{UAMT} proposed UA-MT for left atrium segmentation, which exploits the uncertainty of the teacher model by filtering out unreliable predictions and providing only confident voxel-level predictions for student. Sedai et al.~\cite{Sedai} adaptively weighted regions with uncertain soft labels to guide the student model in OCT images task. These methods rely on a manually set threshold to control the information flow from teacher to student. Thus, they are incapable of tackling incorrect prediction with low uncertainty and likely filtrate valuable guidance for student by mistake. %To address these problem, we directly utilize uncertainty to motivate a better prediction from teacher model.

In this paper, we propose a double-uncertainty weighted method for semi-supervised learning, using a better captured uncertainty to make the teaching-learning process accurate and reliable. The teacher model is trained using Monte-Carlo dropout~\cite{MC} as an approximation of Bayesian Neural Network (BNN) to obtain a double-uncertainty, which consists of segmentation uncertainty and feature one. The former is based on the model's prediction output, while the latter captures uncertainty information existing in each convolution kernel when they detect and extract features. Based on uncertainty estimations and model's predictions, a learnable uncertainty consistency loss is proposed to modify teacher's predictions in an interactive manner. Total loss is the weighted sum of supervised segmentation loss trained from labeled data and consistency loss on both labeled and unlabeled data. This balance is crucial~\cite{Pseudo}, but little-noticed in existing semi-supervised approaches. We take the double-uncertainty of each prediction into consideration, namely assigning smaller loss weight to trustless results.

To the best of our knowledge, we are the first to explore and evaluate feature uncertainty, which reveals the internal mechanism of a model's decision making. Without ground-truth, our newly designed consistency loss still incentivizes teacher's prediction closer to target probabilities and reduces prediction uncertainties. Furthermore, our designed uncertainty weight effectively benefits from unlabeled data and avoids uncertain prediction disturbing supervised training. We conduct exhaustive experiments on the datasets of $2018$ Atrial Segmentation Challenge\footnote{http://atriaseg2018.cardiacatlas.org/data/} for left atrium(LA) segmentation and MICCAI $2019$ KiTS Challenge\footnote{https://kits19.grand-challenge.org/data/} for kidney segmentation. Our method outperforms the state-of-the-art uncertainty-based semi-supervised segmentation approaches.
%and further proposed a double uncertainty-weighted approach to incentivized the model to produce more confident estimates.
\begin{figure}[htb]
		\centering
		\includegraphics[width=4.7in]{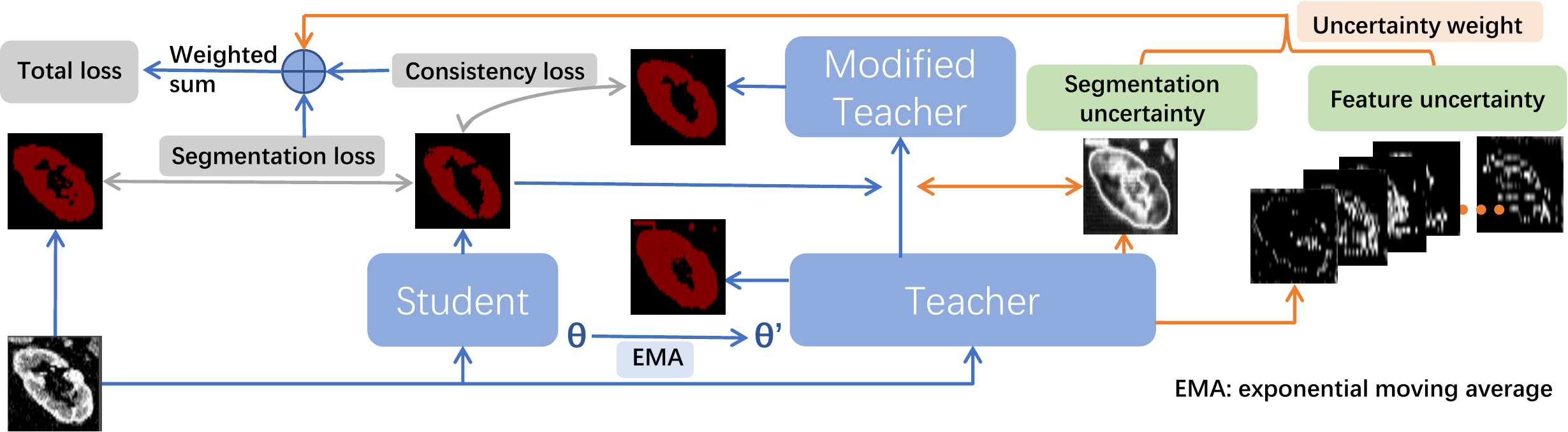}
	\caption{Overview of our model. Teacher model utilizes EMA weights of the student model and generates predictions and double-uncertainty estimates. Uncertainty modifies the teacher in an interactive manner and serves as a weight. Student learns from the teacher by minimizing the weighted sum of the segmentation loss computed from labeled data and consistency loss on both labeled and unlabeled data. }
	\label{fig:network}
\end{figure}
\section{Method}
In this section, we first explore how feature uncertainty exists in channels and propose an approach to evaluating it. 
Then, we detail the method that exploits the feature and segmentation uncertainties into our semi-supervised model. Specifically, we propose a learnable uncertainty consistency loss to form more accurate and less uncertain predictions from the teacher. We further discuss how this double-uncertainty can be interpreted as consistency loss weight to benefit unsupervised training. The whole framework is shown in Fig.~\ref{fig:network}.
\subsection{Feature uncertainty}
\label{sec_feature_uncertainty}
Uncertainty information can be excavated not only from prediction outputs, but also feature extracting process of convolution kernel. 
%We explore this uncertainty through channels and produce feature uncertainty map. 
Each channel of a feature map can be considered as a feature detector focusing on `what' and `where' is an informative part. %As the network training is to adjust the weight parameters for channel aggregating, it is hard to judge the quality of channels. 
If a convolution kernel can extract meaningful features, then the activated regions in its corresponding channel should be basically consistent under multiple inference operations for the same input. Therefore, the uncertainty of each channel implies their learning capability separately. Furthermore, the uncertainty of a feature map can indicate the performance of a model on each given input.
Following~\cite{MC}, we train our model with dropout and random noise at $T$ inference times. For each input, we obtain $T$ intermediate feature maps $F^{t} \in \mathbb{R}^{H \times W \times C}$ at the same layer, which can be considered as a set of $H \times W$ dimensional vectors of the corresponding channels $c \in \{1...C\}$. The uncertainty of channel $c$ is defined by:
\begin{equation}
u_c=\sigma\big(\sum_{p} \sum_{q} g\left(x_{c}^{p}, x_{c}^{q}\right)\big) \quad p, q \in\{1 \ldots T\},
\end{equation}
\begin{equation}
 U_c=mean\big(\sum_{h,w}u_c^{h,w}\big) \quad h\in\{1 \ldots H\}, w\in\{1 \ldots W\},
 \label{eq_Uf}
\end{equation}
where $x_{c}^{p}$ and $x_{c}^{q}$ are two $H \times W$ dimensional vectors of the same channel $c$ at different inference time $p$ and $q$. Function $g(x,y)$ computes an absolute value of difference between these two vectors. Through min-max normalization operation $\sigma$, each channel's uncertainty map $u_c\in \mathbb{R}^{H \times W}$ can be estimated, and uncertainty value $U_c$ is obtained by Eqn. \ref{eq_Uf}. This relative uncertainty among channels represents their information capture ability. Higher uncertainty values imply flatter distributions over extracted channel's features. 
Accordingly, the feature uncertainty of the model at this layer can be obtained by:    
\begin{equation}
	U_{f}=\frac{1}{C} \sum_{i}^{C}\left(U_{i}-\min U_{c}\right).
\end{equation}
This equation calculates an overall estimation of the model's feature extraction ability. %Sec.~\ref{expri:feature_uncertainty} shows feature uncertainty maps and its training process.
%shows the feature uncertainty dropped with enough trained on similar data and indicates the alleviate of feature redundancies. It is therefore a trainable process to minimize the feature uncertainty by adjusting each channel's weight parameters. 
%相对不确定性：同一输入每个channel的uncertainty；
%同一个网络对于同一个任务，不同的输入，每一个channel的相对不确定性是基本固定的
%样本特征不确定性：不同输入，同一层feature map的uncertainty
%通过训练，样本不确定性是不断减小的，
\subsection{Double-uncertainty weighted model}
%In this subsection, we will introduce how to effectively employ both feature uncertainty and segmentation uncertainty in our semi-supervised task. 
Following~\cite{UAMT}, we adopt Mean Teacher~\cite{MT} as our framework and baseline V-Net~\cite{vnet} as teacher and student networks. The weights of the student model at training step $t$ are denoted by $\theta_{t}$. The teacher model uses the exponential moving average(EMA) weights of the student model as $\theta_{t}^{\prime}=\alpha \theta_{t-1}^{\prime}+(1-\alpha) \theta_{t}$, where $\alpha$ is a hyper-parameter called EMA decay. It is utilized to update the weights of student model with gradient descent. This ensembling over training steps improves the quality of the predictions. Meanwhile, random noise on the input and dropout in training process of the teacher model serve as training regularization to improve results. The student aims to learn by minimizing the supervised segmentation loss on labeled data and consistency loss on both labeled and unlabeled data. This consistency enforces consistent prediction between the student and teacher model. Inspired by~\cite{OOD}, the teacher model is also considered as an uncertainty estimation branch in parallel with itself as a prediction branch. We design a learnable uncertainty consistency loss to modify teacher's predictions and facilitate the model to produce less uncertain estimates. Meanwhile, uncertainty is utilized as a weight to balance the training between labeled data and unlabeled data, which significantly benefits the optimization process.  
\subsubsection{Learnable uncertainty consistency loss}
Segmentation uncertainty $U_s$ is captured like~\cite{MC} using the following equation:
\begin{equation}
	u_i=\frac{1}{T} \sum_{t=1}^{T} \operatorname{Softmax}\left(\rho_{t}\right) \qquad
u_v=-\sum_{i=1}^{M} u_{i} \log u_{i},
\label{eq_Us}
\end{equation}
where $\rho_{t}$ is the prediction logits of class $i$ at $t^{th}$ times and $M$ is the number of classes which equals to two in our tasks. $u_i$ is the average softmax probability of $T$ times stochastic dropout sampling from teacher model and $u_v$ is uncertainty metric on voxel $v$. The whole segmentation uncertainty $U_s$ for the given input is obtained by the mean value of each voxel's uncertainty estimate. The prediction probabilities of the teacher model are adjusted as $t_{i}^{\prime}$ by interpolating between the original prediction $t_{i}$ and the prediction of student model $s_{i}$ voxel by voxel. 
\begin{equation}
	t_{i}^{\prime}=(1-u_v) t_{i}+u_v s_{i},
\end{equation}
\begin{equation}
	\mathcal{L}_{\mathrm{c}}=-\frac{1}{V}\sum_{v}^{V}(\sum_{i}^{M} \log \left(t_{v,i}^{\prime}\right)s_{v,i} +\beta \log(1-u_{v,i})), 
\end{equation}
where $v$ represents the $v^{th}$ voxel and $\beta$ is a hyper-parameter for the uncertainty log penalty, preventing the teacher model from producing high uncertainty estimates all the time. 
The designed consistency loss $\mathcal{L}_{\mathrm{c}}$ not only represents whether the teacher's prediction is reliable enough for the student on voxel level, but also produces an interesting dynamic process to be optimized. When $u_{v,i} \rightarrow 1$, which means the prediction of teacher on voxel $v$ is extremely untrustworthy, then $t_{i}^{\prime} \rightarrow s_{i}$, narrows the distance between two models and mitigates teacher's guidance. Conversely, if $u_{v,i} \rightarrow 0$, $t_{i}^{\prime} \rightarrow t_{i}$, then teacher's prediction remains unchanged and keeps its trustful guidance. The above interaction mechanism guarantees the consistency loss can be reduced by the teacher model when providing less uncertainty guidance. 
\subsubsection{Uncertainty weight}
The overall loss is a combination of supervised segmentation loss $\mathcal{L_{\mathrm{s}}}$ and proposed consistency loss $\mathcal{L}_{\mathrm{c}}$, calculated by:
\begin{equation}
\label{eq:loss}
\mathcal{L}=\mathcal{L}_{\mathrm{s}}+\lambda \mathcal{L}_{\mathrm{c}}.
\end{equation}
Proper scheduling of consistency loss weight $\lambda$ is very important~\cite{Pseudo}. Most of the existing choice of $\lambda$ is a time-dependent Gaussian weighting function $\omega(t)=0.1 * e^{\left(-5\left(1-S / L\right)^{2}\right)}$, where $S$ and $L$ represent the current training step and ramp-up length separately~\cite{Cui}. However, this trade-off ignores the proportion and characteristics of unlabeled data, which is incapable of making the most of the teacher model for guidance. Therefore, we take uncertainty estimates into consideration. 
\begin{equation}
	\lambda=\frac{\omega(t)}{U_{f}} \log \frac{1}{U_{s}}=-\frac{\omega(t)}{U_{f}} \log U_{s},
\end{equation}
where $U_{f}$ and $U_s$ represent feature uncertainty and segmentation uncertainty. It is noted that if batch size is bigger than one, $U_f$ and $U_s$ are the average of each input in a mini-batch. While $\omega(t)$ grasps overall share of consistency loss to ramp up from zero along a Gaussian curve, this double-uncertainty serves as weight priors to scale loss targeting on each prediction. Given an input with large uncertainty value $U_{f}$ and $U_s$ predicted by teacher model, $\lambda$ drops in case of wrong guidance disturbing network training. To avoid poor local minima due to extremely small $U_s$, we use a $log$ function to restrict its value. Double-uncertainty controls training loss together, leading to a convincing weighted result.%Since the weight utilized feature uncertainty on channel level, not just the segmentation uncertainty after prediction, the unlabeled data will have better uncertainty representation, leading to convincing weight balance result.
\section{Experiment}
The proposed method is evaluated on the datasets of 2018 Atrial Segmentation Challenge for left atrium (LA) segmentation and MICCAI 2019 KiTS Challenge for kidney segmentation. We compared our method with advanced supervised methods and state-of-the-art semi-supervised methods separately. We adopt Dice, Jaccard, the $95\%$ Hausdorff Distance ($95$HD) and the Average Surface Distance (ASD) as our assessment metrics.
\subsection{Implementation details}
The implementation is based on Pytorch using an NVIDIA Tesla V100 $32$GB GPU. The model is trained using the SGD optimizer and a batch size of $4$ with a gradually decaying learning rate of $0.01$, which is divided by $10$ after each 2500 training steps. Stochastic dropout with $p=0.5$ is applied to layers of the encoder and decoder for $T=16$ times. The supervised segmentation loss is a summation of Cross Entropy Loss and Dice Loss. The $4^{th}$ upsampling layer with $32$ channels is adopted for feature uncertainty estimation. EMA decay $\alpha$ is set as $0.99$, referring to \cite{MT,UAMT}. Hyperparameters $\beta$ controls a log penalty and just a small value can prevent the teacher model from producing high uncertainty estimates. We test various values and $0.001$ is adopted. 
%放一个表格对比1:4
%放一个曲线图收敛速度
\begin{table}\footnotesize
	\caption{Comparison with advanced supervised and semi-supervised methods.}\label{comparison} 
	\setlength{\tabcolsep}{2mm}{
		\begin{tabular}{ccccccc}
			\toprule  %添加表格头部粗线
			\multirow{1}*{Dataset} & \multirow{1}*{Labeled}  & \multirow{1}*{Method} % & \multicolumn{4}{c}{Metrics}   \\
			%\cline{4-7}  % 这部分是画一条横线在2-6 排之间
			& Dice & Jaccard & 95HD & ASD \\
			\midrule  %添加表格中横线
			\multirow{5}*{LA}
			&80 & Baseline V-Net~\cite{vnet} & 0.9025 &
			0.8240 & 8.29 & 1.91 \\
			\cline{2-7}  %添加表格中横线
			& \multirow{2}*{16} & UA-MT~\cite{UAMT} & 0.8888 &
			0.8021 & 7.32 & 2.26 \\
			&& Ours & \textbf{0.8965} & \textbf{0.8135} & \textbf{7.04} &\textbf{2.03}\\
			\cline{2-7}  %添加表格中横线
			&\multirow{2}*{8} & UA-MT~\cite{UAMT} & 0.8425 & 0.7348 & 13.83 & 3.36\\
			&& Ours & \textbf{0.8591} & \textbf{0.7575} & \textbf{12.67} & \textbf{3.31}\\
			\midrule  %添加表格中横线
			
			\multirow{4}*{Kidney}
			& \multirow{4}*{4} 
			& Baseline V-Net~\cite{vnet} & 0.8173 &	0.7291 & 8.90 & 2.75 \\
			%&& nnUnet~\cite{nnUnet} & 0.8614 & 0.8039 & 12.26 & 3.49\\
			&& MT-Sedai et al.~\cite{Sedai} & 0.8593 & 0.7248& 9.26 & 3.06 \\
			&& UA-MT~\cite{UAMT} & 0.8713 &	0.7866 & 11.74 & 3.56 \\
			&& Ours & \textbf{0.8879} & \textbf{0.8169} & \textbf{8.04} &\textbf{2.34}\\
			\bottomrule %添加表格底部粗线
		\end{tabular}
		\footnotesize
	}
\end{table}
\subsubsection{LA segmentation}
This dataset includes $100$ 3D gadolinium-enhanced magnetic resonance imaging scans (GE-MRIs) with segmentation masks. We select $80$ samples as a training set, and the rest $20$ data for testing. For better comparison, we use the same data preprocessing method as~\cite{UAMT}. Table~\ref{comparison} shows under $16$ and  $8$ labeled data, our method outperforms the SOTA semi-supervised method  UA-MT~\cite{UAMT} on the four index. Note that our method trained with only $16$ labeled data performs close to baseline V-Net trained with all $80$ labeled data.% The above comparison indicates the effectiveness and robustness of our semi-supervised method.
\subsubsection{Kidney segmentation}
\begin{figure}[htbp]
	\centering
	\subfigure[]{
		\begin{minipage}[t]{0.32\linewidth}%0.48 2.3in
			%\centering
			\includegraphics[width=1.7in]{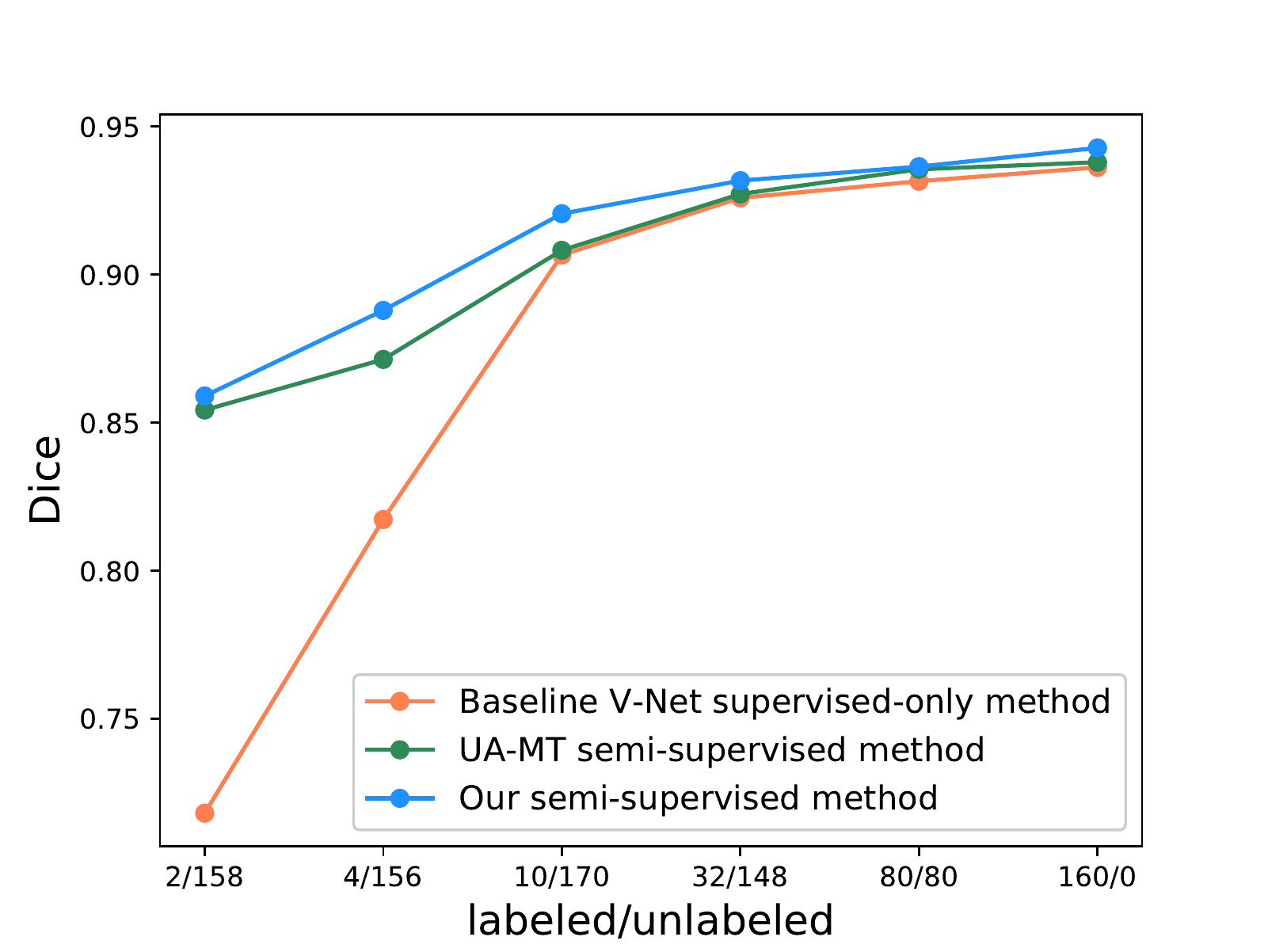}
			\label{kidney}
		\end{minipage}
	}%
	\subfigure[]{
		\begin{minipage}[t]{0.32\linewidth}%0.49
			\includegraphics[width=1.7in]{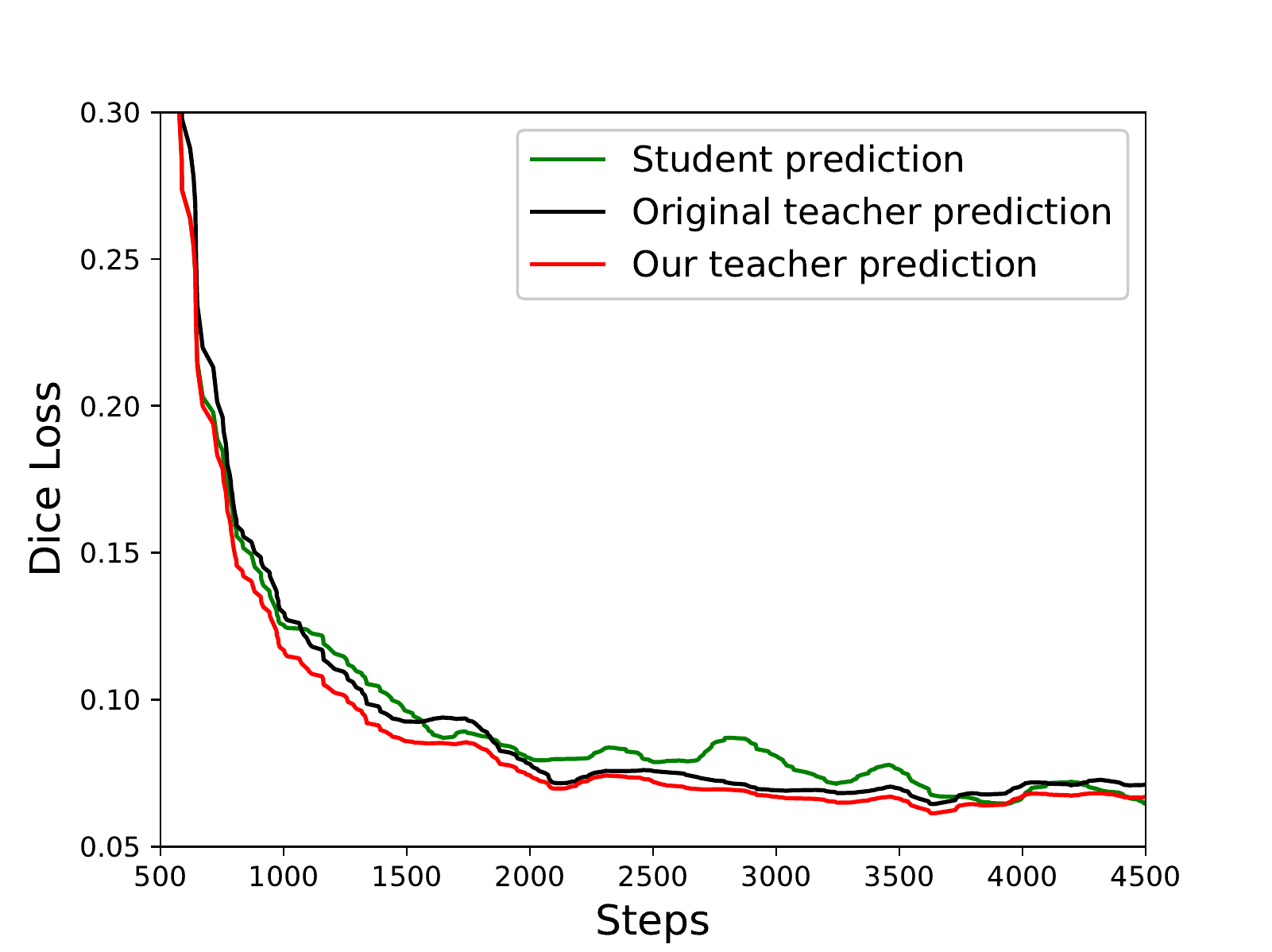}
			\label{new_te}
		\end{minipage}
	}%
	\subfigure[]{
		\begin{minipage}[t]{0.34\linewidth}
			\includegraphics[width=1.7in]{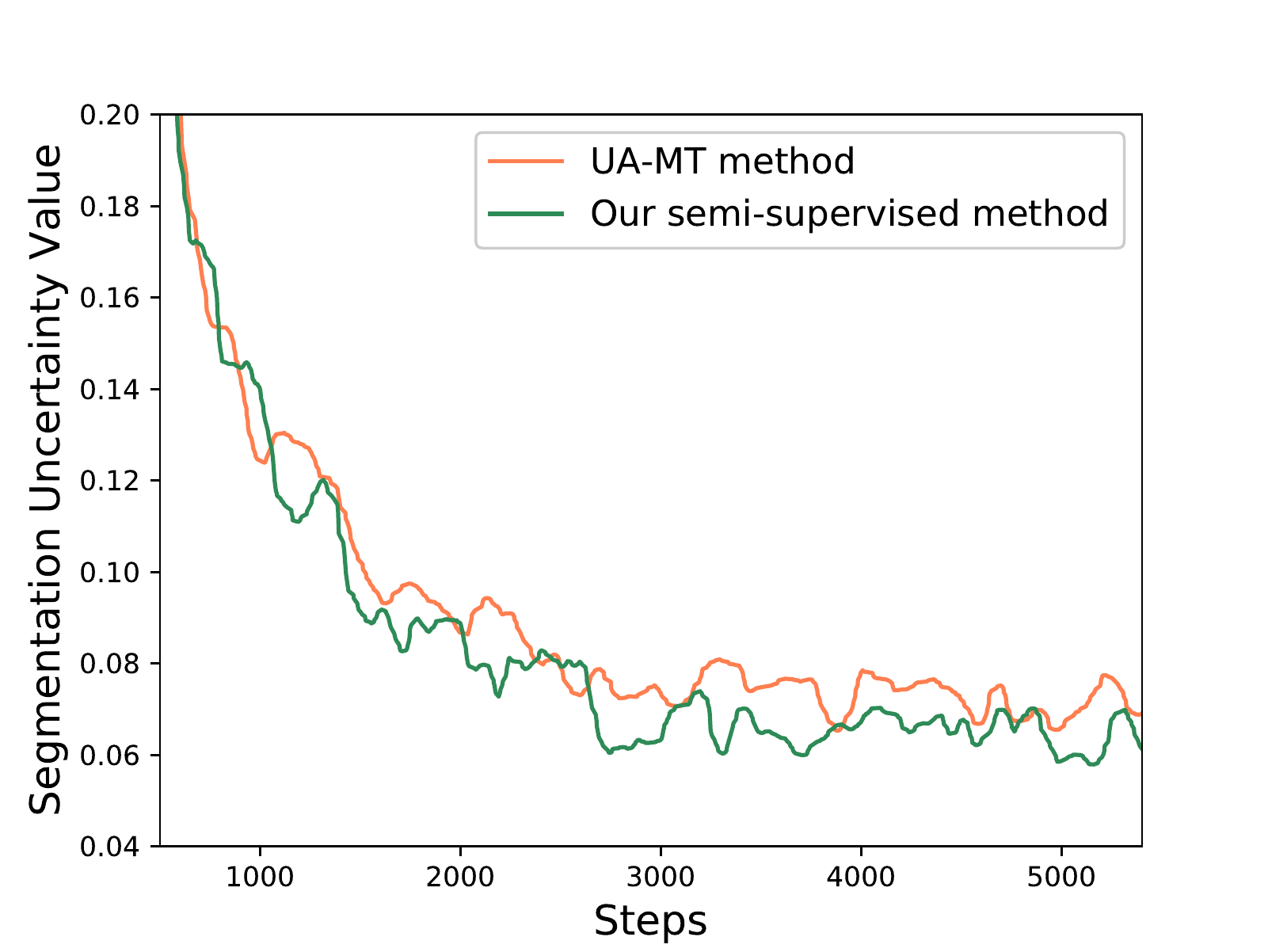}
			\label{la_uncertainty}
		\end{minipage}
	}
	\centering
	\caption{(a) Results of kidney segmentation with different labeled/unlabeled data. (b) Smoothened Dice loss of training data for our modified teacher model, original teacher model and student model. (c) Mean value of segmentation uncertainty from teacher model after each training step.}
\end{figure}

%放一个表格对比1：79 1:4  baselinevnet UAMT Sedai Ours
%放一个曲线图，画出baseline unet，UAMT 和 Ours的dice指标变化， 2,4,10，32,80,160个labeled的对比
This challenging dataset is a collection of contrast-enhanced $3$D abdominal CT scans along with their segmentation ground-truth. We split the $210$ given scans into $160$ for training and $50$ for testing. We extract $3$D patches centering at the kidney region.
To make the results more convincing, a comprehensive comparison with existing methods is conducted. %For supervised-only methods, we select baseline V-Net~\cite{vnet}, and nnUnet~\cite{nnUnet}, which is proved to be the most efficient method submitted to the KiTS 2019 challenge leaderboard. 
Baseline V-Net~\cite{vnet} is selected as a supervised-only method. For semi-supervised methods, besides UA-MT, we choose another latest uncertainty guided semi-supervised method~\cite{Sedai}, which achieves great success in the OCT image task. For a fair comparison, we re-implement this method with the same Mean Teacher architecture, which is referred to as MT-Sedai et al. The same V-Net is adopted as a Bayesian network to estimate uncertainty in all these methods. Table~\ref{comparison} shows our method achieves a high accuracy using only $4$ labeled scans, which ranks top among all compared methods. From Dice comparison between the supervised-only method and semi-supervised methods under different labeled/unlabeled scans in Fig.~\ref{kidney}, it is observed that our method improves performance significantly, especially when the amount of labeled data is small.  %We also compare Dice score under different labeled/unlabeled scans in Fig.~\ref{kidney}. It shows that comparing with supervised-only method, semi-supervised methods improve performance significantly and our method performs better, especially when the amount of labeled data is small. %When we add more labeled scans, results tend to be consistent. 
%\makeatletter\def\@captype{figure}\makeatother
%\begin{minipage}[t]{0.32\linewidth}%0.48 2.3in
%	%\centering
%	\includegraphics[width=1.5in]{img/kidney.eps}
%	\caption{Results on kidney images with different labeled/unlabeled data}
%	\label{kidney}
%\end{minipage}
%\makeatletter\def\@captype{figure}\makeatother
%\begin{minipage}[t]{0.32\linewidth}%0.49
%	\includegraphics[width=1.5in]{img/la_uncertainty.eps}
%	\label{la_uncertainty}
%	\caption{Mean segmentation uncertainty of teacher model after each training step}
%\end{minipage}
%\makeatletter\def\@captype{figure}\makeatother
%\begin{minipage}[t]{0.32\linewidth}
%	\includegraphics[width=1.5in]{img/la_newte.eps}
%	\label{la_uncertainty2}
%	\caption{Smoothened Dice loss of training data}
%\end{minipage}
\subsection{Uncertainty validation}
\label{expri:feature_uncertainty}
\begin{figure}[htbp]
	\centering
	\subfigure{
		\begin{minipage}[t]{0.55\linewidth}
			\centering
			\includegraphics[width=2.2in]{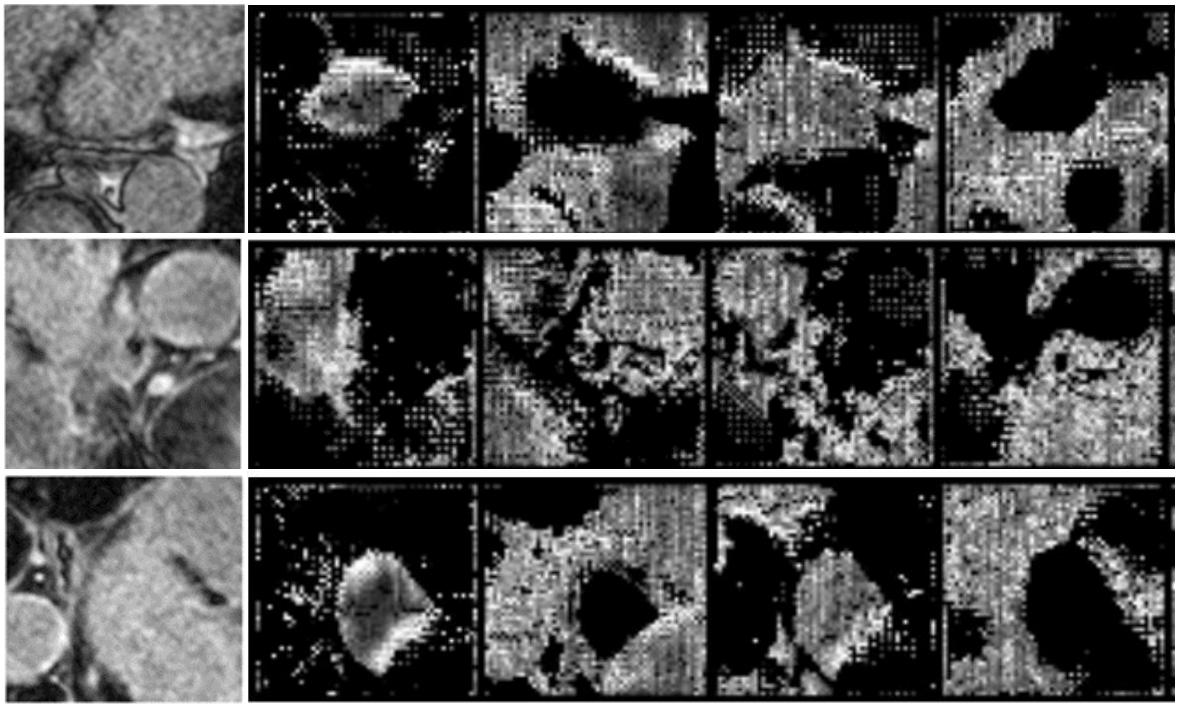}
			%\caption{fig1}
			\label{f_unc_vis}
		\end{minipage}%
	}%
	\subfigure{
		\begin{minipage}[t]{0.45\linewidth}
			\centering
			\includegraphics[width=1.8in]{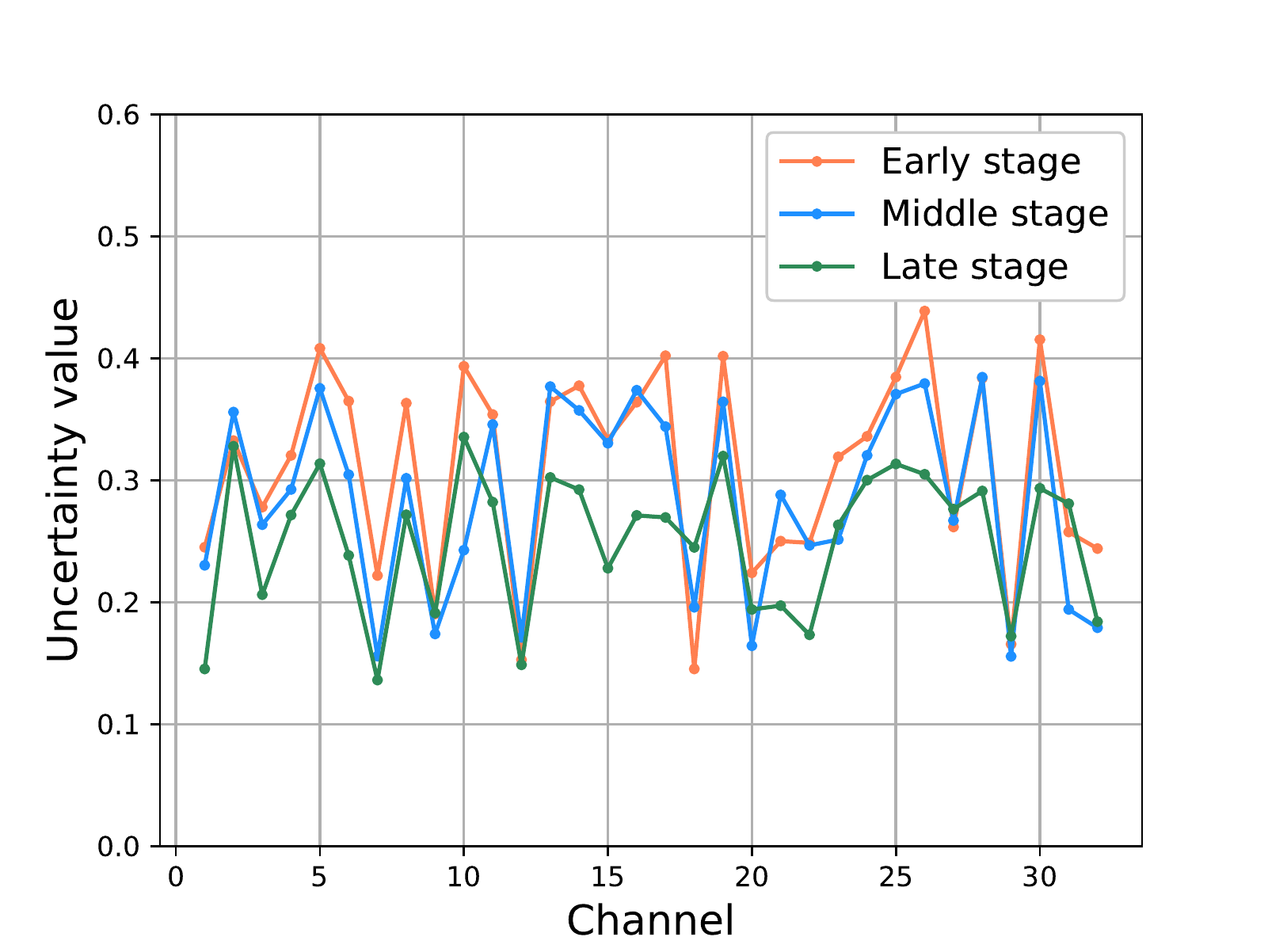}
			%\caption{fig2}
			\label{f_unc_32}
		\end{minipage}%
	}%
	\centering
	\caption{Feature uncertainty. Left: Three input images from early, middle, late stage and their first four channels' feature uncertainty maps (brighter regions represent more uncertainty). Right: Feature uncertainty value of 32 channels.}
	\label{feature_uncertainty}
\end{figure}

Fig.~\ref{feature_uncertainty} shows an example of LA segmentation task under $16$ labeled data.  Experiments show the choice of layers to obtain feature uncertainty does not affect the segmentation results above. Thus, we choose $4^{th}$ upsampling layer for better visualization of uncertainty maps. We randomly select three samples separately from the early, middle and late training stages and calculate their feature uncertainty. Fig.~\ref{feature_uncertainty}(left) compares feature uncertainty maps of their first four channels ($32$ channels in total). Taking the early stage sample as an example, four channels' corresponding uncertainty values are $[0.2453, 0.3327, 0.2782, 0.3206]$. Obviously, the first channel has a relatively better ability to capture the target's features. From feature uncertainty map, uncertainty regions are mainly in the vicinity of the target, especially at its boundary and tissue joint. Conversely, the second channel is proved to be highly uncertain. Its uncertainty map indicates its corresponding convolution kernel is not sure about what and where to see. This provides a way to analyze channels' quality. Fig.~\ref{feature_uncertainty}(right) shows all the uncertainty value of their $32$ channels. It can be seen that 1) relative uncertainty among channels remains basically consistent. 2) as trained, feature uncertainty decreases (These three samples have diminishing feature uncertainty values of $[0.3109,0.2838,0.2565]$, using Eqn. \ref{eq_Uf}). 

Consistency loss defines the prediction inconsistency between student and teacher model. Thus, to investigate the ability of our designed consistency loss equals to verify our modification to the teacher model. By calculating Dice Loss of training data with the ground-truth label, Fig.~\ref{new_te} shows our modified teacher achieves better predictions.
%Fig.~\ref{la_te} shows how teacher model can be effectively enhanced. By calculating dice loss with groundtruth label, we compare the prediction of student model, original teacher model and our uncertainty modified teacher model with training data. It is worth noting that though most training data are unlabeled, our modified teacher can always lead to better prediction. Even in the early stages when the teacher model can not achieve good performance, our modified teacher can still provide a valuable guidance for student model. 
A more accurate teacher speeds up the feedback loop, leading to better test performance, as shown in Table~\ref{ablation}. Additionally, the interaction between uncertainty estimates and model's predictions in our proposed loss function facilitates teachers to produce less uncertain results, which is proven in Fig.~\ref{la_uncertainty}.
We also validate the proposed uncertainty weight in Table~\ref{ablation}. Particularly, with uncertainty weight, UA-MT~\cite{UAMT} is improved as well.
\begin{table}\footnotesize
	\caption{Contributions of proposed consistency loss and uncertainty weight.}
	\label{ablation}
	\setlength{\tabcolsep}{2mm}{
		\begin{tabular}{lcc}
			\toprule
			Method  & 10 labeled & 32 labeled\\
			\midrule  %添加表格中横线
			MSE loss~\cite{MT}  & 0.8944 & 0.9215\\
			UA-MT loss~\cite{UAMT} & 0.9081 & 0.9252\\
			UA-MT loss~\cite{UAMT} + Weight & 0.9098 & 0.9346\\
			\midrule  %添加表格中横线
			Our Consistency loss & 0.9182 & 0.9358\\
			Our Consistency loss + Weight & \textbf{0.9206} & \textbf{0.9365}\\
			\bottomrule
		\end{tabular}\footnotesize}
\end{table}
\section{Conclusion}
In this paper, we make novel contributions to employ uncertainty estimates to the teacher-student model for semi-supervised learning. It is the first time that feature uncertainty is proposed and evaluated. Our learnable uncertainty consistency loss facilitates a more accurate teacher model with less uncertainty. Uncertainty is also utilized as a consistency weight to balance unsupervised training. Comprehensive experimental analysis on two medical datasets shows the significant improvements of our method using limited labeled images. %Future work focuses on exploiting feature uncertainty and alleviating model's uncertainty to enhance segmentation performance.
\bibliographystyle{splncs04}

%\bibliography{paper647}

\begin{thebibliography}{10}
\providecommand{\url}[1]{\texttt{#1}}
\providecommand{\urlprefix}{URL }
\providecommand{\doi}[1]{https://doi.org/#1}

\bibitem{Bai2017}
Bai, W., Oktay, O., Sinclair, M., Suzuki, H., Rajchl, M., Tarroni, G., Glocker,
  B., King, A.P., Matthews, P.M., Rueckert, D.: Semi-supervised learning for
  network-based cardiac {MR} image segmentation. In: Medical Image Computing
  and Computer Assisted Intervention - {MICCAI} 2017 - 20th International
  Conference, Quebec City, QC, Canada, September 11-13, 2017, Proceedings, Part
  {II}. pp. 253--260 (2017). \doi{10.1007/978-3-319-66185-8\_29},
  \url{https://doi.org/10.1007/978-3-319-66185-8\_29}

\bibitem{Cui}
Cui, W., Liu, Y., Li, Y., Guo, M., Li, Y., Li, X., Wang, T., Zeng, X., Ye, C.:
  Semi-supervised brain lesion segmentation with an adapted mean teacher model.
  In: Information Processing in Medical Imaging - 26th International
  Conference, {IPMI} 2019, Hong Kong, China, June 2-7, 2019, Proceedings. pp.
  554--565 (2019). \doi{10.1007/978-3-030-20351-1\_43},
  \url{https://doi.org/10.1007/978-3-030-20351-1\_43}

\bibitem{OOD}
DeVries, T., Taylor, G.W.: Learning confidence for out-of-distribution
  detection in neural networks. CoRR  \textbf{abs/1802.04865} (2018),
  \url{http://arxiv.org/abs/1802.04865}

\bibitem{MC}
Gal, Y., Ghahramani, Z.: Dropout as a bayesian approximation: Representing
  model uncertainty in deep learning. In: Proceedings of the 33nd International
  Conference on Machine Learning, {ICML} 2016, New York City, NY, USA, June
  19-24, 2016. pp. 1050--1059 (2016),
  \url{http://proceedings.mlr.press/v48/gal16.html}

\bibitem{Temporal}
Laine, S., Aila, T.: Temporal ensembling for semi-supervised learning. In: 5th
  International Conference on Learning Representations, {ICLR} 2017, Toulon,
  France, April 24-26, 2017, Conference Track Proceedings (2017),
  \url{https://openreview.net/forum?id=BJ6oOfqge}

\bibitem{Ensemble}
Lakshminarayanan, B., Pritzel, A., Blundell, C.: Simple and scalable predictive
  uncertainty estimation using deep ensembles. In: Advances in Neural
  Information Processing Systems 30: Annual Conference on Neural Information
  Processing Systems 2017, 4-9 December 2017, Long Beach, CA, {USA}. pp.
  6402--6413 (2017),
  \url{http://papers.nips.cc/paper/7219-simple-and-scalable-predictive-uncertainty-estimation-using-deep-ensembles}

\bibitem{Pseudo}
Lee, D.H.: Pseudo-label : The simple and efficient semi-supervised learning
  method for deep neural networks. ICML 2013 Workshop : Challenges in
  Representation Learning (WREPL)  (07 2013)

\bibitem{trainsformation}
Li, X., Yu, L., Chen, H., Fu, C., Heng, P.: Semi-supervised skin lesion
  segmentation via transformation consistent self-ensembling model. In: British
  Machine Vision Conference 2018, {BMVC} 2018, Northumbria University,
  Newcastle, UK, September 3-6, 2018. p.~63 (2018),
  \url{http://bmvc2018.org/contents/papers/0162.pdf}

\bibitem{vnet}
Milletari, F., Navab, N., Ahmadi, S.: V-net: Fully convolutional neural
  networks for volumetric medical image segmentation. In: Fourth International
  Conference on 3D Vision, 3DV 2016, Stanford, CA, USA, October 25-28, 2016.
  pp. 565--571 (2016). \doi{10.1109/3DV.2016.79},
  \url{https://doi.org/10.1109/3DV.2016.79}

\bibitem{Few-shot}
Mondal, A.K., Dolz, J., Desrosiers, C.: Few-shot 3d multi-modal medical image
  segmentation using generative adversarial learning. CoRR
  \textbf{abs/1810.12241} (2018), \url{http://arxiv.org/abs/1810.12241}

\bibitem{unet}
Ronneberger, O., Fischer, P., Brox, T.: U-net: Convolutional networks for
  biomedical image segmentation. In: Medical Image Computing and
  Computer-Assisted Intervention - {MICCAI} 2015 - 18th International
  Conference Munich, Germany, October 5 - 9, 2015, Proceedings, Part {III}. pp.
  234--241 (2015). \doi{10.1007/978-3-319-24574-4\_28},
  \url{https://doi.org/10.1007/978-3-319-24574-4\_28}

\bibitem{Sedai}
Sedai, S., Antony, B.J., Rai, R., Jones, K., Ishikawa, H., Schuman, J.S.,
  Wollstein, G., Garnavi, R.: Uncertainty guided semi-supervised segmentation
  of retinal layers in {OCT} images. In: Medical Image Computing and Computer
  Assisted Intervention - {MICCAI} 2019 - 22nd International Conference,
  Shenzhen, China, October 13-17, 2019, Proceedings, Part {I}. pp. 282--290
  (2019). \doi{10.1007/978-3-030-32239-7\_32},
  \url{https://doi.org/10.1007/978-3-030-32239-7\_32}

\bibitem{MT}
Tarvainen, A., Valpola, H.: Mean teachers are better role models:
  Weight-averaged consistency targets improve semi-supervised deep learning
  results. In: Advances in Neural Information Processing Systems 30: Annual
  Conference on Neural Information Processing Systems 2017, 4-9 December 2017,
  Long Beach, CA, {USA}. pp. 1195--1204 (2017)

\bibitem{UAMT}
Yu, L., Wang, S., Li, X., Fu, C., Heng, P.: Uncertainty-aware self-ensembling
  model for semi-supervised 3d left atrium segmentation. In: Medical Image
  Computing and Computer Assisted Intervention - {MICCAI} 2019 - 22nd
  International Conference, Shenzhen, China, October 13-17, 2019, Proceedings,
  Part {II}. pp. 605--613 (2019). \doi{10.1007/978-3-030-32245-8\_67},
  \url{https://doi.org/10.1007/978-3-030-32245-8\_67}

\bibitem{DAN}
Zhang, Y., Yang, L., Chen, J., Fredericksen, M., Hughes, D.P., Chen, D.Z.: Deep
  adversarial networks for biomedical image segmentation utilizing unannotated
  images. In: Medical Image Computing and Computer Assisted Intervention -
  {MICCAI} 2017 - 20th International Conference, Quebec City, QC, Canada,
  September 11-13, 2017, Proceedings, Part {III}. pp. 408--416 (2017).
  \doi{10.1007/978-3-319-66179-7\_47},
  \url{https://doi.org/10.1007/978-3-319-66179-7\_47}

\end{thebibliography}
%
\end{document}